\pdfoutput=1

\documentclass[11pt]{article}

\usepackage[preprint]{acl}

\usepackage{times}
\usepackage{latexsym}
\usepackage{array}
\usepackage{array}
\usepackage[T1]{fontenc}

\usepackage[utf8]{inputenc}

\usepackage{microtype}

\usepackage{inconsolata}

\usepackage{graphicx}

\usepackage{inconsolata}
\usepackage{color}
\usepackage[noend]{algpseudocode}
\usepackage{bm}
\usepackage{algorithmicx,algorithm}
\usepackage{multirow}
\usepackage{subcaption}
\usepackage{amsfonts}
\usepackage{amsmath}
\usepackage{booktabs}
\usepackage{makecell}
\usepackage{graphicx}

\usepackage{colortbl}

\usepackage{CJKutf8}

\title{Thinker-DDM: Modeling Deliberation for Machine Translation with a Drift-Diffusion Process}

\author{
    Hongbin Na$^{\spadesuit,}$\thanks{Equal contribution.}, 
    Zimu Wang$^{\heartsuit,}$\footnotemark[1],
    Mieradilijiang Maimaiti$^{\clubsuit,}$\thanks{Corresponding author.},
    Tong Chen$^\heartsuit$, \\
    \textbf{Wei Wang$^\heartsuit$, Tao Shen$^\spadesuit$, Ling Chen$^\spadesuit$} \\
    $^{\spadesuit}$Australian AI Institute, University of Technology Sydney  \\
    $^{\heartsuit}$Xi'an Jiaotong-Liverpool University $\quad$
    $^\clubsuit$Chinese Academy of Sciences $\quad$
    \\
    \texttt{Hongbin.Na@student.uts.edu.au, \{Zimu.Wang19,Tong.Chen19\}@student.xjtlu.edu.cn} \\
    \texttt{miradel\_51@hotmail.com, Wei.Wang03@xjtlu.edu.cn} \\
    \texttt{\{Tao.Shen,Ling.Chen\}@uts.edu.au}
}

\newcommand \footnoteONLYtext[1]
{
	\let \mybackup \thefootnote
	\let \thefootnote \relax
	\footnotetext{#1}
	\let \thefootnote \mybackup
	\let \mybackup \imareallyundefinedcommand
}

\begin{document}
\begin{CJK}{UTF8}{gbsn}
\maketitle

\begin{abstract}
Large language models (LLMs) have demonstrated promising potential in various downstream tasks, including machine translation. However, prior work on LLM-based machine translation has mainly focused on better utilizing training data, demonstrations, or pre-defined and universal knowledge to improve performance, with a lack of consideration of decision-making like human translators. In this paper, we incorporate \textbf{Thinker} with the \textbf{D}rift-\textbf{D}iffusion \textbf{M}odel (Thinker-DDM) to address this issue. We then redefine the Drift-Diffusion process to emulate human translators' dynamic decision-making under constrained resources. We conduct extensive experiments under the high-resource, low-resource, and commonsense translation settings using the WMT22 and CommonMT datasets, in which Thinker-DDM outperforms baselines in the first two scenarios. We also perform additional analysis and evaluation on commonsense translation to illustrate the high effectiveness and efficacy of the proposed method.
\end{abstract}

\section{Introduction}

Large language models (LLMs), such as GPT-4 \cite{OpenAI-2023-GPT-4}, GLM-130B \cite{Zeng-etal-2023-GLM}, and LLaMA \cite{Touvron-etal-2023-LLaMA}, have recently achieved the state-of-the-art performance in a variety of downstream tasks, such as information extraction \cite{Peng-etal-2023-ICL}, text summarization \cite{Wang-etal-2023-Summarization}, and adversarial attacks \cite{Wang-etal-2023-LLM-Attack}. One particular area where LLMs have demonstrated promising potential is machine translation \cite{Zhang-etal-2022-OPT,Vilar-etal-2023-PaLMMT}, and they also excel under low-resource and zero-resource scenarios \cite{Moslem-etal-2023-AdaptiveMT,Zhu-etal-2023-MultilingualMT}.

\begin{figure}[t!]
    \centering
    \includegraphics[width=1\linewidth]{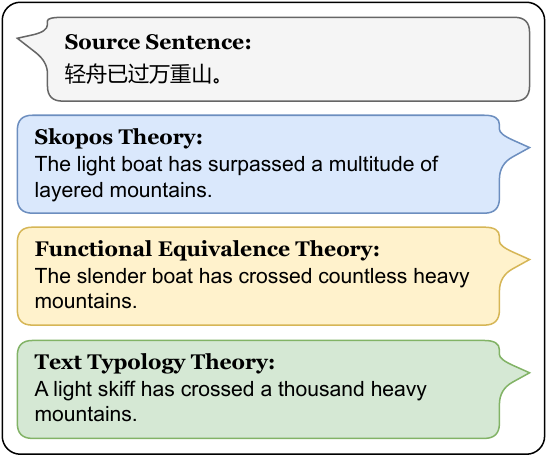}
    \caption{An example of translating a Chinese poetry to English using different translation theories. \textbf{Skopos theory} emphasizes the objectives (mood and emotion); \textbf{functional equivalence theory} underscores target culture; \textbf{text typology theory} concentrates on the characteristics of different text types (poetic and rhythmic).}
    \label{fig:translation-ex}
\end{figure}

Recent research on LLM-based machine translation can be broadly categorized into two paradigms: in-context learning (ICL, \citealp{Brown-etal-2020-GPT-3}) that conditions on natural language instructions or demonstrations, and fine-tuning that updates the model parameters based on the availability of usually limited amount of labelled data. In the context of ICL, researchers concentrate on leveraging optimal in-context examples \cite{Agrawal-etal-2023-ICExamples,Sarti-etal-2023-RAMP,Iyer-etal-2023-Disambiguation}, dictionary knowledge \cite{Ghazvininejad-etal-2023-Dictionary,Lu-etal-2023-Dictionary}, adaptive learning \cite{Moslem-etal-2023-AdaptiveMT,Reinauer-etal-2023-Adaptive}, and translation memories \cite{Reheman-etal-2023-Memories}. Numerous studies have also fine-tuned LLMs to augment their capacity in translating unseen languages \cite{Yang-etal-2023-BigTranslate,Mao-etal-2024-LowResource} and domains \cite{Moslem-etal-2022-DomainSpecific,Moslem-etal-2023-DomainSpecific} and building multilingual LLMs \cite{Zhang-etal-2023-BayLing,Zhu-etal-2023-NonEnglish}.

Despite this success, prior work on LLM-based machine translation, whether through in-context learning or fine-tuning, shares a common limitation: they typically apply a uniform translation strategy \cite{He-etal-2023-MAPS} to all sentences within a given task. This static approach contrasts sharply with the methodology of human translators. A human expert engages in a dynamic and adaptive process, deliberating on the unique challenges posed by each sentence. For instance, they may consider various established translation strategies, such as the Skopos theory \cite{reiss2010grundlegung}, functional equivalence theory \cite{nida1964toward}, and text typology theory \cite{reiss1989text} (see Figure \ref{fig:translation-ex}), and apply them interchangeably based on the source text's nuances. The crucial challenge, therefore, is not merely providing the model with access to different strategies, but equipping it with a mechanism to decide when and how to apply them, mimicking the resource-constrained deliberation of a human. This cognitive aspect of translation—the sentence-level, dynamic decision-making—remains largely unexplored.

We introduce a new machine translation approach that incorporates \textbf{Thinker} with \textbf{D}rift-\textbf{D}iffusion \textbf{M}odel (Thinker-DDM) to address the aforementioned challenges. Thinker-DDM incorporates a decision-making process characterized by drift and diffusion to simulate human cognitive behavior. Specifically, we first design relevant translation strategy prompts in line with the three representative theories in translation. Then, we redefine the Drift-Diffusion process, incorporating the processes of initial bias (drift), evidence gathering (diffusion), and boundary-driven decision-making to emulate how human translators behave with resource constraints. We conduct extensive experiments under the high-resource, low-resource, and commonsense translation settings using the WMT22 \cite{kocmi-etal-2022-findings} and CommonMT \cite{he-etal-2020-box} datasets. Evaluation results show that Thinker-DDM outperforms Microsoft Translator, GPT-3.5, and Hybrid Max-Routing \cite{hendy2023good} baselines in the first two scenarios. We also carry out additional analysis and evaluation on the translation strategies and DDM, which highlight the effectiveness and efficacy of the proposed method.


The main contributions are summarized as follows: (1) We propose Thinker-DDM, a novel machine translation approach to simulate human translators’ dynamic decision-making process. (2) We define relevant translation strategy prompts in line with representative theories in translation and redefine the Drift-Diffusion process to adapt to the machine translation task. (3) We conduct comprehensive experiments across multiple languages to demonstrate the high effectiveness and efficacy of the proposed method.

\section{Preliminaries}

\paragraph{Translation Studies.} Translation studies is a multidisciplinary field that centers on translation as a means of cross-cultural and cross-linguistic communication \cite{munday2022introducing}. Theorists have developed a variety of theories to guide and interpret the translation process, encompassing significant schools of thought and perspectives. As shown in Figure \ref{fig:translation-ex}, \emph{Skopos theory} \cite{reiss2010grundlegung} posits that translation activities should revolve around their pre-determined objectives, emphasizing the guiding role of these purposes in shaping translation strategies. Complementing this, \emph{functional equivalence theory} \cite{nida1964toward} underscores the function and impact of translations within the target culture, focusing on how translations bridge meaning across diverse cultures. Additionally, \emph{text typology theory} \cite{reiss1989text} concentrates on the unique characteristics of different text types and their influence on translation requirements, advocating for adaptive translation strategies tailored to specific text categories, such as informative, expressive, or operative. Together, these theories construct a profound perspective that transcends literal translation, highlighting the importance of cultural adaptability, target audience needs, and effective communication in the translation process \cite{IJESL4440}.

\paragraph{Individual Decision-Making Processes.} In exploring the nuances of individual decision-making, dual-system theory offers a key framework, differentiating two modes of thinking: intuitive System 1 and analytical System 2. System 1 governs low-risk, familiar decisions with its fast, automated processes, whereas System 2 is engaged in complex, logical decision-making scenarios \cite{kahneman2011thinking}. Despite its insights, this theory has limitations in fully capturing decision-making dynamics. Here, DDM presents a more nuanced understanding by portraying decision-making as a continuous accumulation of evidence, leading to a choice \cite{Ratcliff1978ATO}. It highlights the process's stability and speed, especially in decisions grounded in clear value orientations, aligning with value-based decision theory \cite{Rowland1946TheoryOG}, elucidating how internal values and external information guide rapid, precise choices \cite{Ratcliff2016DiffusionDM}.

\section{Methodology}

\begin{figure*}[t]
    \centering
    \includegraphics[width=1\linewidth]{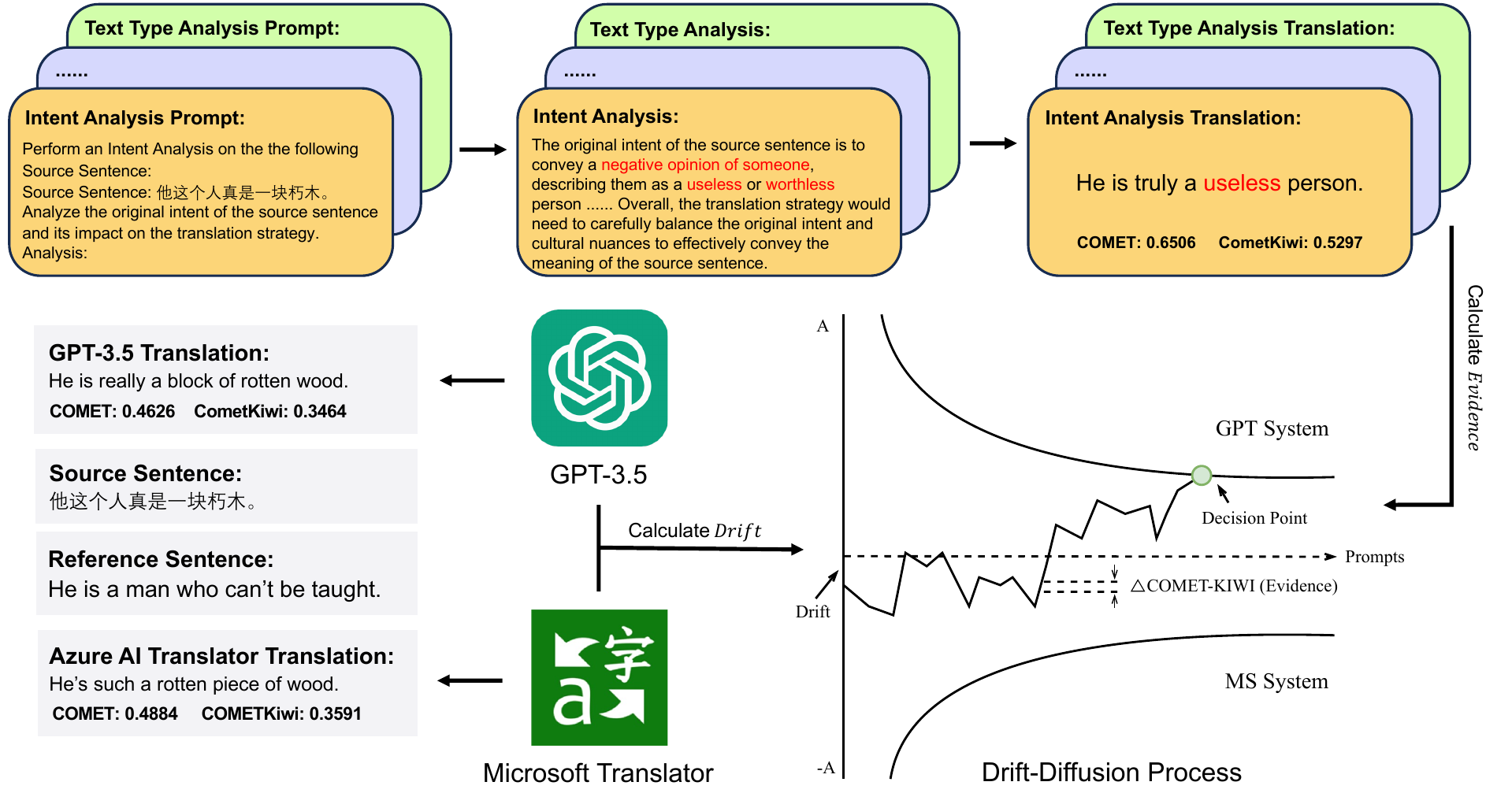}
    \caption{An overview of our proposed Thinker-DDM method. It first utilizes relevant translation strategy prompts in line with the three representative theories in translation, and then redefine the Drift-Diffusion process, incorporating the processes of initial bias (drift), evidence gathering (diffusion), and boundary-driven decision-making.}
    \label{fig:method}
\end{figure*}

Following the overall framework of our proposed Thinker-DDM method illustrated in Figure \ref{fig:method}, in this section, we introduce each part of the framework in detail.

\subsection{Translation Strategy Prompts}
\begin{figure}[t!]
    \centering
    \includegraphics[width=1\linewidth]{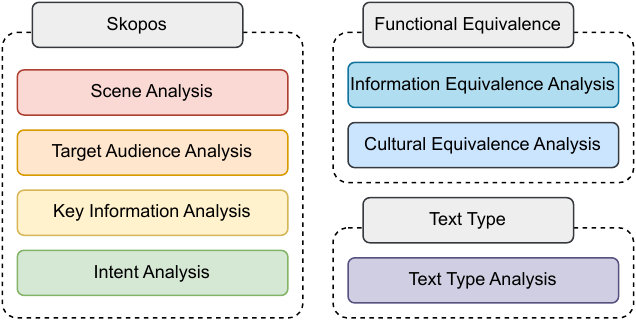}
    \caption{Our suggested translation strategies for Thinker-DDM and their underlying theoritical frameworks.}
    \label{fig:strategies}
\end{figure}

In the pursuit of optimizing translation quality, we first develop a series of translation strategy prompts that draw on representative theories in translation studies. These prompts facilitate a deeper analysis of source sentences, taking into account various aspects during human translation, such as target audiences, key information, information equivalence, and culture equivalence. This ensures a satisfactory level of linguistic accuracy as well as cultural and contextual relevance. Figure \ref{fig:strategies} shows our suggested translation strategies, which are categorized according to their underlying theoretical framework：Skopos theory, functional equivalence theory, and text type theory. With each prompt category targeting a specific facet of the translation process, it is possible to conduct a thorough analysis that will guide the translation strategy effectively. The detailed content and examples of each of the seven prompts are organized in Appendix \ref{sec:prompts}.

This structured approach, grounded in well-established theoretical frameworks, offers a comprehensive methodology to address the multifaceted nature of translation. Systematically applying these prompts to source sentences enriches the translation process and guarantees that translations are accurate linguistically as well as appropriately attuned to contextual and cultural nuances. While the translation strategy prompts provide valuable insights for source sentences, they may also introduce extraneous noises, which lowers the overall quality and accuracy of the translation.

\subsection{Drift-Diffusion Model}
DDM is a framework for quantifying the process of rapid decision-making under uncertainty. It decomposes the decision-making process into several essential components: \textbf{drift}, \textbf{diffusion}, and \textbf{boundary conditions}, as shown in Figure \ref{fig:method}. This model is particularly prevalent in studies involving reaction time and decision accuracy, offering a mathematical approach to understanding cognitive processes \cite{bogacz2006physics}.


\paragraph{Drift} represents the average rate of information accumulation in favor of a particular decision. It is defined as the mean of the stochastic process and is denoted by \(\mu\). Mathematically, it can be expressed as:
\begin{equation}
    \mu = \frac{dE[X(t)]}{dt},
\end{equation}
where \(E[X(t)]\) is the expected value of the decision variable \(X\) at time \(t\). Drift rate is a critical parameter as it reflects the strength of evidence and influences the speed-accuracy tradeoff in the decision-making process.

\paragraph{Diffusion} refers to the stochastic or random component of the decision-making process, which accounts for the inherent variability and unpredictability in accumulating evidence. This random fluctuation is often modeled mathematically as a process with a variance parameter \(\sigma^2\), encapsulating the element of uncertainty in evidence accumulation. The diffusion component of the decision process can be expressed as:
\begin{equation}
    dX(t) = \mu dt + \sigma dB(t),
\end{equation}
where \(dX(t)\) represents the instantaneous change in the decision variable at time \(t\), quantifying the accumulated evidence and \(dB(t)\) denotes the incremental, stochastic change, analogous to the random noise, in the evidence accumulation over time. This term captures the essence of uncertainty and randomness in the decision-making process, reflecting how real-world decisions are influenced by factors that cannot always be predicted or quantified precisely.

\paragraph{Boundary Conditions} define the thresholds for making a decision. When the accumulated evidence, represented by the decision variable \(X(t)\), reaches one of these boundaries, a decision is made. The boundaries are often set symmetrically at \(\pm A\), with \(A\) being a positive constant. These boundaries not only determine the decision outcome but also play a crucial role in modeling reaction times. The time at which \(X(t)\) first hits either boundary is the predicted reaction time for the decision.

\subsection{Translation Thinker with DDM}

In this study, we redefine and apply DDM to translation evaluation, aiming to develop an innovative model for choosing the optimal translation of a given source sentence. It involves a series of iterative steps to determine the optimal translation for a specific sentence, as the overall procedure illustrated in Algorithm \ref{algo}.

\begin{algorithm}[t!]
\caption{Translation Thinker with DDM}
\begin{algorithmic}[1]
\State $\textbf{Input:}$ Original sentence $S$
\State $\textbf{Output:}$ Best translation $T_{best}$
\State Initialize $T_{MS}(S), T_{GPT}(S), \mathcal{P}$;
\State Initialize $A_{up}, A_{low}$$, \text{decay}$;
\State $\textit{Drift} \gets \mathcal{C}(T_{GPT}(S)) - \mathcal{C}(T_{MS}(S))$;
\For{each prompt $P_i$ in $\mathcal{P}$}
    \State Generate new translation $T_{P_i}(S)$;
    \State $\textit{Diffusion}_i \gets \mathcal{C}(T_{P_i}(S)) - \mathcal{C}(T_{MS}(S))$;
    \State Update $\textit{Drift} \gets \textit{Drift} + \textit{Diffusion}_i$;
    \State Update boundaries: 
    \State $A_{up/low} \gets A_{up/low} \times e^{-\text{decay}}$;
    \If{$\textit{Drift} \geq A_{up}$ or $\textit{Drift} \le A_{low}$}
        \State \textbf{break};
    \EndIf
\EndFor
\If{$\textit{Drift} \geq A_{up}$}
    \State $T_{best} \gets$ translation with highest $\mathcal{C}$;
\ElsIf{$A_{up} > \textit{Drift} > A_{low}$}
    \State $T_{best} \gets$ translation with highest $\mathcal{C}$;
\ElsIf{$\textit{Drift} \le  A_{low}$}
    \State $T_{best} \gets T_{MS}(S)$;
\EndIf
\State \textbf{return} $T_{best}$
\end{algorithmic}
\label{algo}
\end{algorithm}

We redefine \textbf{\textit{drift}} as an initial preference, implying that there is a bias towards a particular translation system prior to gathering evidence. We utilize the CometKiwi model $\mathcal{C}$ \cite{rei-etal-2022-cometkiwi} to conduct a reference-free quality assessment to imitate the human translation process. We assess the translation by both Microsoft Translator\footnote{\url{https://www.microsoft.com/en-us/translator/}} $T_{MS}(S)$ and GPT-3.5 $T_{GPT}(S)$, where $S$ is the original sentence. Mathematically, drift is defined as the difference in the CometKiwi scores between these two methods of translation:
\begin{equation}
    \textit{Drift} = \mathcal{C}(T_{GPT}(S)) - \mathcal{C}(T_{MS}(S)).
\end{equation}

Subsequently, a series of translation strategy prompts $\mathcal{P} = \{P_1, \ldots P_i, \ldots, P_n\}$ are proposed to generate new translation versions $T_{P_i}(S)$. The order of cues is randomized to simulate the stochastic process of diffusion. For each prompt $P_i$, its corresponding \textbf{\textit{diffusion}} is calculated as:
\begin{equation}
    \textit{Diffusion}_i = \mathcal{C}(T_{P_i}(S)) - \mathcal{C}(T_{MS}(S)).
\end{equation}
in which $\textit{Diffusion}_i$ represents the evidence collected in each iteration. In each iteration, the drift value $\textit{Drift}$ is updated to $\textit{Drift} + \textit{Diffusion}_i$, while checking whether the pre-determined decision boundaries, the upper boundary $A_{up}$ and the lower boundary $A_{low}$, have been reached. The boundary values are dynamically adjusted and decreases in an exponential decay manner, formally,
\begin{equation}
    A_{up/low}^{new} = A_{up/low}^{old} \times e^{-\text{decay}}.
\end{equation}
Such exponential decay of the boundary values allows the model to maintain a wider decision range in the initial stages, allowing for ample exploration; as time progresses, the boundary values decrease, enabling the model to make decisions faster, thereby enhancing efficiency.

Based on the final value of the $\textit{Drift}$, we select the optimal translation. If $\textit{Drift} \geq A_{up}$, we choose the translation with the highest CometKiwi score from all generated translations; otherwise, if $\textit{Drift} \le A_{low}$, we select Microsoft Translator's result $T_{MS}(S)$. For those that do not reach the boundary until the end of the iteration, the approach is consistent with reaching the upper boundary, and we select the translation with the highest CometKiwi score since all translation candidates have been generated.

\section{Experiments}

\subsection{Datasets}

We conducted experiments on the WMT22 dataset \cite{kocmi-etal-2022-findings}, which is widely utilized in machine translation research. It encompasses both high-resource and low-resource languages, predominantly featuring English (EN) $\Leftrightarrow$ Chinese (ZH), English (EN) $\Leftrightarrow $ German (DE), English (EN) $\Leftrightarrow $ Japanese (JA), and German (DE) $\Leftrightarrow$ French (FR) language pairs. Additionally, to explore translation performance in low-resource languages, particular emphasis of the dataset was placed on Czech (CS) $\Leftrightarrow$ Ukrainian (UK) and English (EN) $\Leftrightarrow$ Ukrainian (UK). We sampled a subset of 500 sentences for each language pair from the test set of the dataset for experiments. 

To further investigate the capability of translation models in handling semantic ambiguities, we incorporated Chinese-to-English translation samples from the CommonMT dataset \cite{he-etal-2020-box}.

\subsection{Evaluation Metrics}

We adopted two mainstream reference-based evaluation metrics in machine translation, COMET \cite{rei-etal-2022-comet} and BLEURT \cite{sellam-etal-2020-bleurt}, to assess Thinker-DDM's performance. These model-based metrics have shown to be superior to conventional string-based metrics, such as BLEU \cite{papineni-etal-2002-bleu}, and have been widely employed in LLM-based machine translation literature \cite{hendy2023good,Moslem-etal-2023-AdaptiveMT}. In accordance with the previous work, we utilized \texttt{wmt22-comet-da}\footnote{\url{https://github.com/Unbabel/COMET}} and \texttt{BLEURT-20}\footnote{\url{https://github.com/google-research/bleurt}} checkpoints for the selected metrics, respectively.

\subsection{Baselines}

We examined Thinker-DDM against a variety of representative baselines, encompassing both the single-candidate and multiple-candidate approaches:

\begin{itemize}
    \item \textbf{Microsoft Translator (MS-Translator)} is an off-the-shelf commercial machine translation system provided by Microsoft that is able to translate text instantly or in batches across more than 100 languages. We accessed the translator through the public API provided by Microsoft Azure\footnote{\url{https://azure.microsoft.com/en-us/products/ai-services/ai-translator/}}.
    \item \textbf{GPT-3.5} is the standard zero-shot translator using the GPT-3.5 model. We employed the prompts as shown in Table \ref{tab:translate-prompt} to ensure precise and cohesive translations.
    \item \textbf{Hybrid Max-Routing (Max-Routing)} determines the upper limit by choosing the superior translation from the two aforementioned systems, i.e. MS-Translator and GPT-3.5, based on CometKiwi evaluations. We referred to it as the ``Hybrid Max-Routing'' approach as illustrated in \newcite{hendy2023good}.
\end{itemize}

\subsection{Experimental Settings}


Thinker-DDM iteratively selected the optimal translation by calculating the new evidence provided by new translation candidates, for which the results were obtained by the GPT-3.5 model. We accessed GPT-3.5 by calling the official API released by OpenAI\footnote{\url{https://platform.openai.com/}}, and we selected the \texttt{gpt-3.5-turbo-instruct} checkpoint, a refined version of GPT-3 designed to perform natural language tasks with heightened accuracy and reduced toxicity. We set the temperature as 0.5 to simulate the randomness in human reasoning. Given that both systems already possess high translation capabilities, we set the initial boundaries $A_{up}$ and $A_{low}$ as 0.05 and -0.05, respectively, and we set the decay as 0.2.

\subsection{Main Results}

\begin{table*}[t!]
\resizebox{\textwidth}{!}{\begin{tabular}{c|ccccccccc}
\toprule

\textbf{Method} & \textbf{DE-EN} & \textbf{EN-DE} & \textbf{ZH-EN} & \textbf{EN-ZH} & \textbf{JA-EN} & \textbf{EN-JA} & \textbf{FR-DE} & \textbf{DE-FR} & \textbf{Average} \\ 
\midrule
               & \multicolumn{9}{c}{\textbf{COMET22}} \\ 
\midrule
\textbf{MS-Translator} & 0.8532 & 0.8716 & 0.8118 & 0.8712 & 0.8234 & 0.8904 & 0.8655 & 0.8429 & 0.8538 \\
\textbf{GPT-3.5} & 0.8527 & 0.8688 & 0.8290 & 0.8663 & 0.8298 & 0.8850 & 0.8689 & 0.8386 & 0.8549 \\ 
\midrule
\textbf{Max-Routing} & 0.8585 & 0.8762 & \underline{0.8299} & \underline{0.8809} & 0.8362 & 0.8969 & \textbf{0.8772} & \underline{0.8478} & 0.8630 \\
\textbf{Thinker-DDM} & \underline{0.8588} & \textbf{0.8765} & \underline{0.8299} & \textbf{0.8810} & \underline{0.8369} & \underline{0.8989} & \underline{0.8757} & \textbf{0.8500} & \underline{0.8635} \\
\textbf{Thinker-ALL} & \textbf{0.8597} & \underline{0.8764} & \textbf{0.8304} & 0.8806 & \textbf{0.8398} & \textbf{0.9012} & 0.8745 & 0.8466 & \textbf{0.8637} \\ 
\midrule
               & \multicolumn{9}{c}{\textbf{BLEURT}} \\ 
\midrule
\textbf{MS-Translator} & 0.7430 & 0.7767 & 0.6898 & 0.7270 & 0.6903 & 0.6959 & 0.7842 & \underline{0.7317} & 0.7298 \\
\textbf{GPT-3.5} & 0.7422 & 0.7726 & 0.7163 & 0.7199 & 0.6959 & 0.6717 & 0.7888 & 0.7119 & 0.7274 \\ 
\midrule
\textbf{Max-Routing} & 0.7495 & \underline{0.7814} & 0.7159 & \textbf{0.7365} & 0.7049 & 0.6997 & \underline{0.7979} & 0.7301 & 0.7395 \\
\textbf{Thinker-DDM} & \textbf{0.7518} & 0.7810 & \underline{0.7173} & \underline{0.7339} & \underline{0.7058} & \underline{0.7029} & \textbf{0.7981} & \textbf{0.7373} & \textbf{0.7410} \\
\textbf{Thinker-ALL} & \underline{0.7513} & \textbf{0.7821} & \textbf{0.7211} & 0.7307 & \textbf{0.7088} & \textbf{0.7034} & 0.7978 & 0.7250 & \underline{0.7400} \\ 
\bottomrule
\end{tabular}}
\caption{High-resource translation results of Thinker-DDM against baselines on the WMT22 dataset, in which the optimal and second-best results are highlighted in \textbf{bold} and \underline{underlined}, respectively.}
\label{tab:main_result}
\end{table*}

\paragraph{High-resource Translation Results} The high-resource translation results of Thinker-DDM against baselines are reported in Table \ref{tab:main_result}. The Thinker-ALL method took all the translation results across different translation strategies as well as MS-Translator and GPT-3.5 into account. It selected the optimal translation results under CometKiwi for the nine candidates. From the table, it is evident that Thinker-DDM and Thinker-ALL exhibited superior translation quality across most evaluated language pairs, outperforming all baselines with both COMET22 and BLEURT metrics. This indicates its effectiveness and generalizability, which is not limited to specific language pairs but is consistent across various linguistic combinations. In addition, there are only marginal differences in scores between Thinker-DDM and Thinker-ALL. Such close performance suggested that Thinker-DDM is highly efficient and stable. Since Thinker-DDM acheived the optimal or second-best results in most cases, it is even more stable than Thinker-ALL. In addition, it potentially reduces computational resources and processing requests.

\paragraph{Low-resource Translation Results}

Machine translation under the low-resource scenario presents a challenge, primarily due to the limited data available for these languages. To investigate the performance of Thinker-DDM under such settings, we conducted experiments and reported the experimental results of translation between low-resource languages (i.e. CS $\Leftrightarrow$ UK) and between high-resource and low-resource languages (i.e. EN $\Leftrightarrow$ UK) under the COMET evaluation metrics in Table \ref{table:LRS}. The results were also consistent with those in high-resource translation. Hybrid Max-Routing, Thinker-DDM, and Thinker-ALL exhibited commendable translation quality, emphasizing the significance of customized and optimized algorithms in the translation of low-resource languages.

\begin{table}[t!]
\resizebox{\columnwidth}{!}{
\begin{tabular}{ccccc}
\toprule
\textbf{Method}             & \textbf{CS-UK}  & \textbf{UK-CS}  & \textbf{EN-UK}  & \textbf{UK-EN}  \\ \midrule
\textbf{MS-Translator}      & 0.9096          & 0.9149          & 0.8842          & 0.8566          \\
\textbf{GPT-3.5}   & 0.9014          & 0.9056          & 0.8746          & 0.8529          \\ \midrule
\textbf{Max-Routing} & 0.9156          & 0.9211          & 0.8898          & \textbf{0.8634} \\
\textbf{Thinker-DDM}        & \textbf{0.9195} & \underline{0.9243}          & \underline{0.8950}          & \underline{0.8630}          \\
\textbf{Thinker-ALL}        & \underline{0.9176}          & \textbf{0.9245} & \textbf{0.8980} & 0.8574          \\ \bottomrule
\end{tabular}
}
\caption{Low-resource translation results of Thinker-DDM against baselines under the COMET evaluation metric.}
\label{table:LRS}
\end{table}

\paragraph{Commonsense Translation Results}
In the field of machine translation, data-driven approaches have become dominant, leading to a notable decline in the explicit exploration of world knowledge\cite{he-etal-2020-box}. Our method attempted to bridge this gap to handle translation tasks requiring a comprehensive grasp of commonsense knowledge and cultural background. However, as shown in Table \ref{Table:CMT}, the performance of Thinker-DDM in commonsense translation did not meet expectations. We hypothesized that this might be related to the inability of the CometKiwi evaluation metric to assess translation quality accurately.

We conducted supplementary experiments to test our hypothesis, wherein the translation quality evaluator, CometKiwi, was substituted with COMET. As shown in Table \ref{Table:CMT}, the experimental results witnessed a substantial enhancement with both COMET and BLEURT metrics, implying that incorporating translation theories can endow models with a more profound comprehension of context and cultural nuances. Contrarily, CometKiwi lacks the capability to discriminate between varying levels of translation quality in commonsense machine translation; this also implies the need in designing a better reference-free evaluation metric in the future.

\begin{table}[t!]
\resizebox{\columnwidth}{!}{
\begin{tabular}{ccc}
\toprule
\textbf{Method}             & \textbf{COMET} & \textbf{BLEURT} \\ \midrule
\textbf{MS-Translator}      & 0.8156           & 0.7010                 \\
\textbf{GPT-3.5}   & 0.8405           & 0.7326                \\ \midrule
\textbf{Max-Routing} & 0.8383           & 0.7278                \\
\textbf{Thinker-DDM}        & 0.8340           & 0.7244               \\
\textbf{Thinker-ALL}        & 0.8364           & 0.7253               \\
\midrule
\textbf{Thinker-DDM (COMET)}       & \underline{0.8739}               & \underline{0.7713}                 \\
\textbf{Thinker-ALL (COMET)}       & \textbf{0.8818}                & \textbf{0.7853}                  \\ \bottomrule
\end{tabular}
}
\caption{Commonsense translation results of Thinker-DDM against baselines.}
\label{Table:CMT}
\end{table}

\subsection{Additional Analysis}

\paragraph{Effectiveness of Translation Strategies}

\begin{figure}[t!]
    \centering
    \includegraphics[width=1\linewidth]{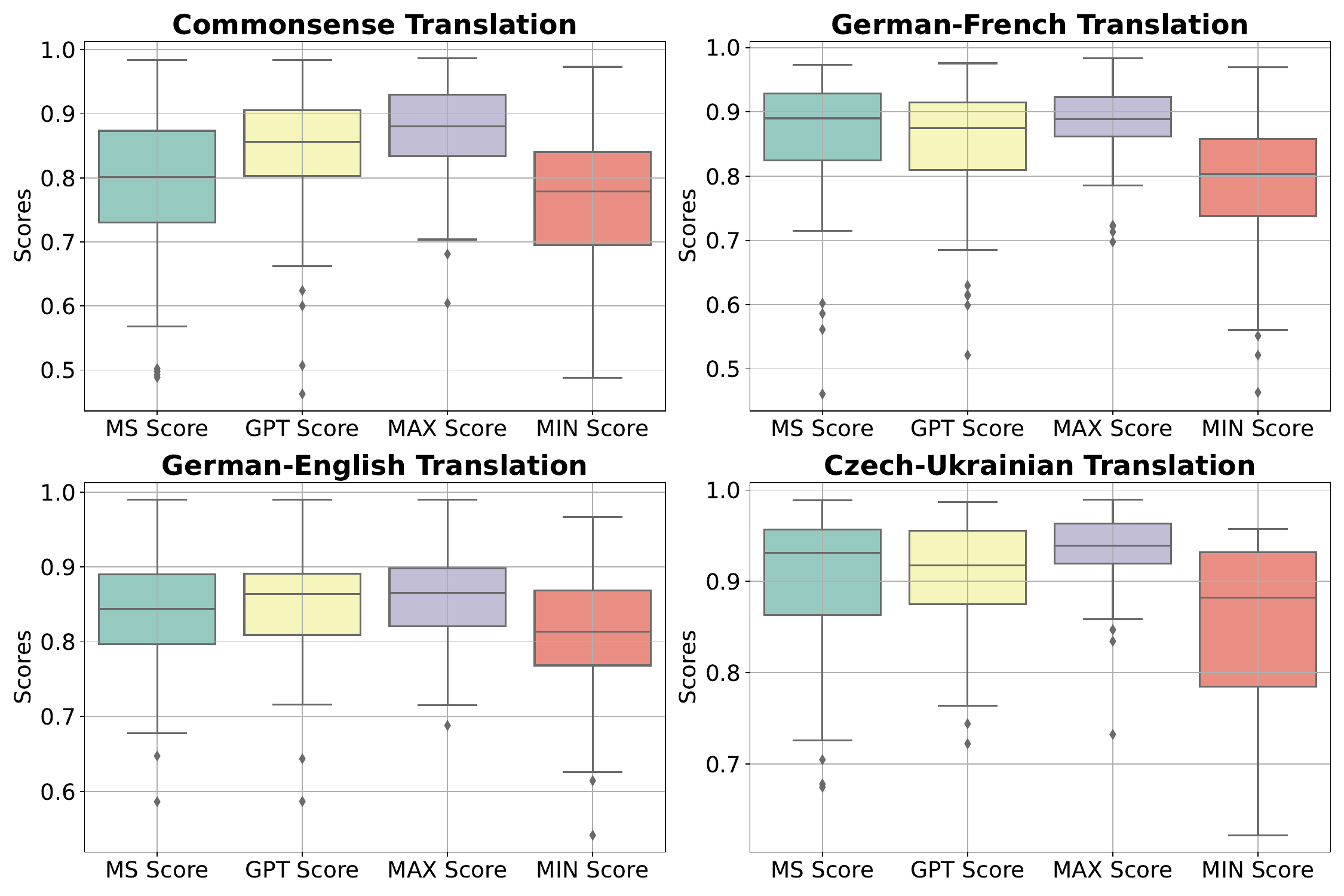}
    \caption{Experimental results of the COMET scores of MS-Translator's and GPT-3.5's translations and the maximum and minimum scores post-prompt.}
    \label{fig:prompts}
\end{figure}

In exploring the impact of translation strategies, our initial focus was on their role in expanding the candidate space. To understand their roles more profoundly, we conducted an additional experiment as follows: we first randomly sampled 50 sentence pairs from commonsense translation, DE-FR, DE-EN, and CS-UK. Then we calculated the scores of MS-Translator and GPT-3.5, as well as the maximum and minimum scores of post-prompt under the COMET metric.

As shown in Figure \ref{fig:prompts}, the translation strategies across all translation settings provided both higher upper and lower bottom limits, indicating the positive role of translation strategies in enhancing translation performance. More importantly, the risks of performance degradation due to the lower limit can be efficiently eliminated through the Drift-Diffusion process. This ensures the quality and stability of translation while expanding the range of candidate translations.

\begin{figure}[t!]
    \centering
    \includegraphics[width=1\linewidth]{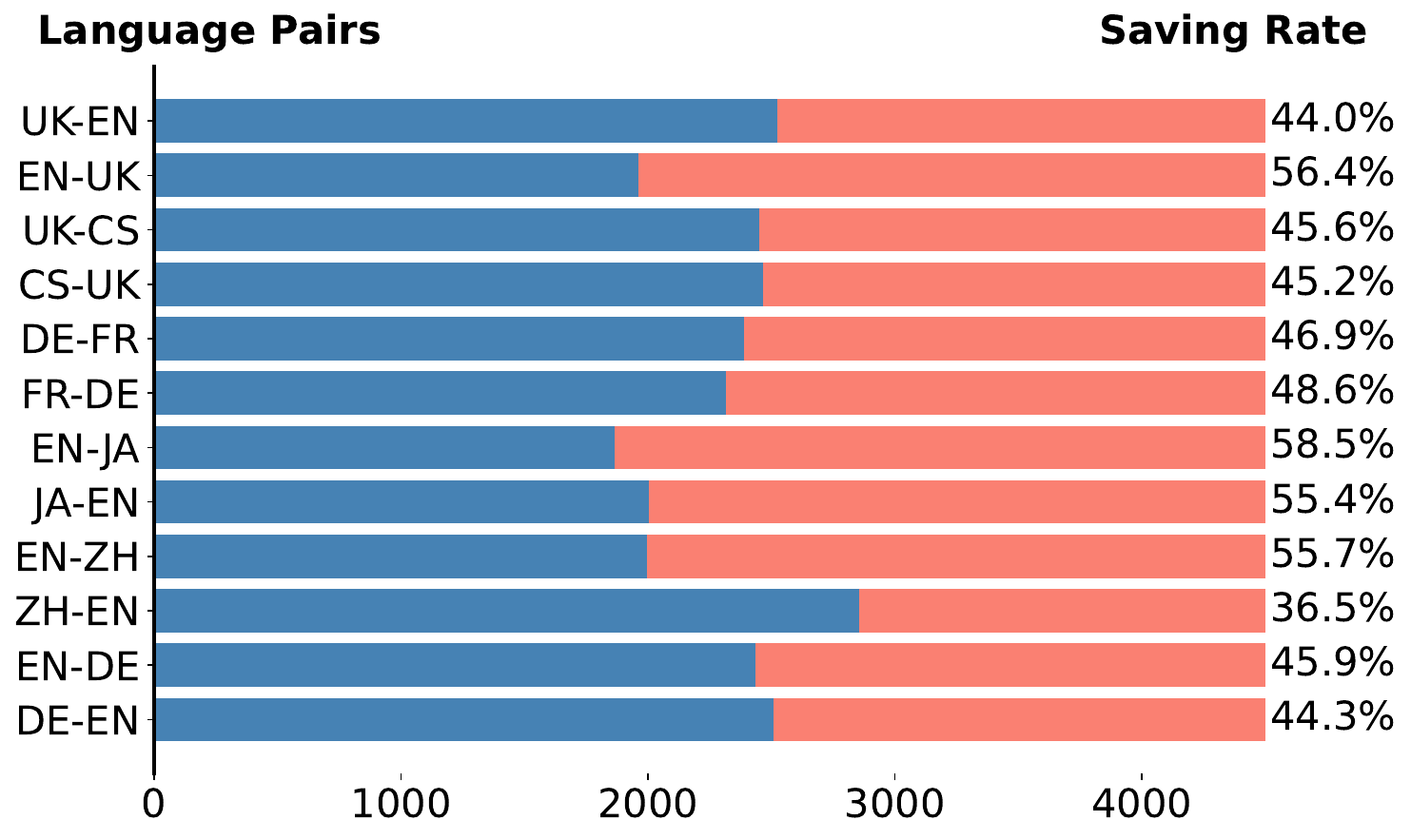}
    \caption{Query saving rate across different language pairs.}
    \label{fig:api-saving}
\end{figure}

\paragraph{Effectiveness of Drift-Diffusion Model}
In the main experiments, we concluded the effectiveness of DDM on translation accuracy; however, the effectiveness of DDM in translation efficiency remains explored. Following the experimental results shown in Figure \ref{fig:api-saving}, Thinker-DDM achieved an average reduction of 48\% in query volume compared to the Thinker-ALL, underscoring our approach's enhanced efficiency and underlining the significance of decision theory optimization in settings with limited resources while achieving satisfactory translation outcomes. 

\begin{figure}[t!]
    \centering
    \includegraphics[width=1\linewidth]{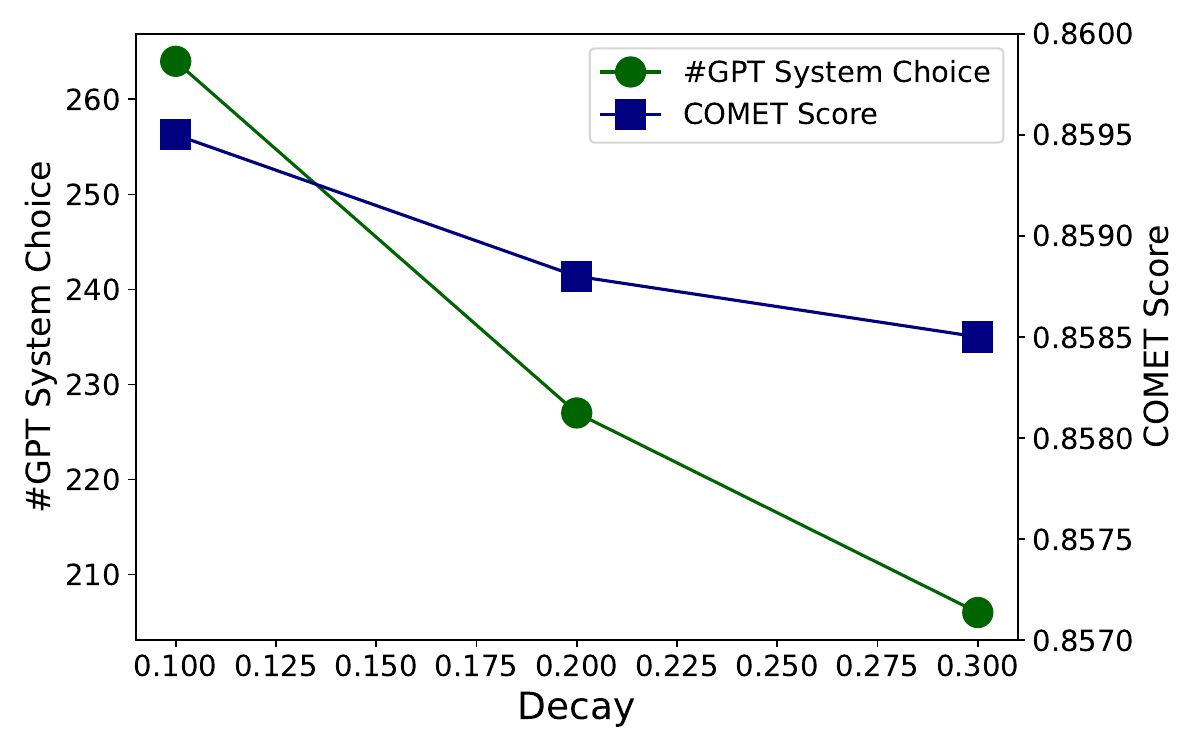}
    \caption{Correlations between the decay factor and the number of GPT system choice and translation performance.}
    \label{fig:decay}
\end{figure}

Building on the aforementioned insights, we finally turned our attention to the influence of the decay factor on decision-making in translation tasks. To this end, we focused on the DE-EN translation, set the boundary at 0.05 and experimented with decay values of \{0.1, 0.2, 0.3\}. As shown in Figure \ref{fig:decay}, the decay value had a negative correlation with the number of selecting the GPT system, indicating the significance of selecting an appropriate decay value for balancing translation effectiveness and efficiency.

\section{Related Work}

\paragraph{Decision Theory in LLMs}

LLMs have recently made significant strides in emulating human cognitive processes, particularly focusing on enhancing decision-making and problem-solving capabilities. System 2 Attention \cite{weston20232} and Tree of Thoughts (ToT, \citealp{yao2023tree}) are exemplary in this regard. Both initiatives mark a shift towards mimicking human ``System 2'' cognitive processes, involving conscious, analytical thought, and logical reasoning.
In a similar vein, developments in agent-based systems have also garnered attention \cite{liu2023llmpowered, tian2023duma, wang2023humanoid}. These systems collectively emphasize the integration of rapid responsiveness, robust reasoning, and emotional in LLMs, showcasing a trend towards creating more interactive, intuitive, and human-like LLMs. Additionally, DOMINO \cite{wang2023domino} presents a dual-system approach for multi-step visual language reasoning, adeptly managing complex questions and information extraction. Together, these advancements illustrate the broadening scope of decision theory in LLMs, moving towards systems that not only mimic human cognitive processes but also interact and reason in dynamically complex environments.

\paragraph{Machine Translation} Machine translation has been extensively studied over the past few years by transforming multiple paradigms. Early research utilizes rule-based methods \cite{Forcada-etal-2011-RuleMT} to handle the machine translation task; some other methods exploit statistical \cite{Koehn-etal-2003-StatisticalMT01,Koehn-etal-2007-StatisticalMT02} and deep learning \cite{zheng-etal-2021-self,9515788} techniques to enhance translation performance.

Recently, the rapid advancement of LLMs catalyzes the research on LLM-based machine translation, consisting of two paradigms: ICL and fine-tuning. In the context of ICL, researchers concentrate on leveraging optimal in-context examples \cite{Agrawal-etal-2023-ICExamples,Sarti-etal-2023-RAMP,Iyer-etal-2023-Disambiguation}, dictionary knowledge \cite{Ghazvininejad-etal-2023-Dictionary,Lu-etal-2023-Dictionary}, adaptive learning \cite{Moslem-etal-2023-AdaptiveMT,Reinauer-etal-2023-Adaptive}, and translation memories \cite{Reheman-etal-2023-Memories}. Additionally, numerous studies have also fine-tuned LLMs to augment their capacity in translating unseen languages \cite{Yang-etal-2023-BigTranslate,Mao-etal-2024-LowResource} and domains \cite{Moslem-etal-2022-DomainSpecific,Moslem-etal-2023-DomainSpecific} and building multilingual LLMs \cite{Zhang-etal-2023-BayLing,Zhu-etal-2023-NonEnglish}. Furthermore, some research also concentrates on post-editing the translation outcomes \cite{Moslem-etal-2023-Postedit01,Raunak-etal-2023-PostEdit02} and leveraging LLMs for machine translation evaluation \cite{Fu-2023-GPTScore,fernandes-etal-2023-devil}.

\section{Conclusion and Future Work}

We introduced Thinker-DDM, a novel machine translation approach that simulates the dynamic decision-making process of human translators. We designed relevant translation strategy prompts in line with the three representative theories in translation and then redefined the Drift-Diffusion process to emulate human translators' decision-making under constrained resources. We conducted extensive experiments to validate the effectiveness and efficacy of our proposed method. In the future, we will incorporate Thinker-DDM into more natural language generation tasks, such as text summarization and question answering.

\section*{Limitations}

We organized the limitations of Thinker-DDM two-fold: (1) Thinker-DDM adopted CometKiwi to evaluate the translation qualities; however, since it is a reference-free evaluation model, there exists some gaps between the evaluation results obtained from CometKiwi and COMET, leaving a potential line in developing a more effective reference-free metric. (2) We only tested the performance of Thinker-DDM on one LLM, GPT-3.5, in our experiments. There remain investigations of Thinker-DDM on other commonly used LLMs, e.g. LLaMA 2 \cite{touvron2023llama} and Mistral \cite{jiang2023mistral}, in the future.

\bibliography{custom}

\appendix
\clearpage
\onecolumn
\section{Translation Strategy Prompts}\label{sec:prompts}

\begin{table*}[h]
    \centering
    \begin{tabular}{p{\linewidth}}
        \toprule
        \rowcolor{Red!15}
        \textsc{Instruction:} \\
        \rowcolor{Red!15}
        Perform a Scene Analysis on the the following Source Sentence: \\
        \rowcolor{Red!15}
        Source Sentence: \verb|{source_sentence}| \\
        \rowcolor{Red!15}
        Analyze the context on the following source sentence and consider how this context impacts the translation strategy. \\
        \rowcolor{Red!15}
        Analysis: \\
        \midrule
        \rowcolor{Red!15}
        \textsc{Example:} \\
        \rowcolor{Red!15}
        I will just arrange a returns label for you now. \\
        \midrule
        \rowcolor{Red!15}
        \textsc{Answer:} \\
        \rowcolor{Red!15}
        The source sentence is spoken by a customer service representative to a customer who needs to return an item. The context of this sentence is a professional and polite interaction between the representative and the customer. The representative is offering to help the customer by arranging a returns label, which indicates that the company has a specific process in place for returns. \textcolor{BrickRed}{\textbf{The use of the word ``just'' suggests that this is a quick and easy task for the representative to complete.}} Overall, the context of this sentence requires a translation strategy that is clear, concise, and professional. The translation should accurately convey the representative's willingness to assist the customer and the company's established process for returns. It may also be important to maintain the polite tone and use of the word ``just'' to convey the ease of the task. Additionally, the translation should be culturally appropriate and considerate of any potential language barriers between the representative and the customer. \\
        \midrule
        \rowcolor{Red!15}
        \textsc{MS-Translator Translation:} \\
        \rowcolor{Red!15}
        我现在\textcolor{BrickRed}{只为您}安排一个退货标签。 \\
        \midrule
        \rowcolor{Red!15}
        \textsc{Analysis-based Translation:} \\
        \rowcolor{Red!15}
        我现在\textcolor{BrickRed}{会为您}安排一个退货标签。 \\
        \bottomrule
    \end{tabular}
    \caption{Instruction and example for conducting scene analysis on a given source sentence.}
    \label{tab:scene-analysis}
\end{table*}

\twocolumn

\begin{table*}[t!]
    \centering
    \begin{tabular}{p{\linewidth}}
        \toprule
        \rowcolor{Orange!15}
        \textsc{Instruction:} \\
        \rowcolor{Orange!15}
        Perform a Target Audience Analysis on the the following Source Sentence: \\
        \rowcolor{Orange!15}
        Source Sentence: \verb|{source_sentence}| \\
        \rowcolor{Orange!15}
        Describe and analyze the characteristics of the target audience, discussing how these characteristics influence the translation. \\
        \rowcolor{Orange!15}
        Analysis: \\
        \midrule
        \rowcolor{Orange!15}
        \textsc{Example:} \\
        \rowcolor{Orange!15}
        For quality purposes this chat interaction will be closed, please remember you can always come back and we will be happy to continue assisting you. \\
        \midrule
        \rowcolor{Orange!15}
        \textsc{Answer:} \\
        \rowcolor{Orange!15}
        The target audience for this source sentence appears to be customers or clients who are engaging in a chat interaction with a customer service representative. These individuals are likely seeking assistance or support with a product or service. \\
        \rowcolor{Orange!15}
        \textcolor{BrickRed}{\textbf{The first characteristic of the target audience is that they are seeking quality service.}} This suggests that they are likely to be detail-oriented and have high expectations for the assistance they receive. As a result, the translation should accurately convey the message of the source sentence and maintain a professional tone. \\
        \rowcolor{Orange!15}
        ... \\
        \rowcolor{Orange!15}
        Overall, the target audience for this source sentence appears to be professional, detail-oriented, tech-savvy, time-conscious, and potentially long-term customers. These characteristics should be considered when translating the sentence to ensure that the message is accurately conveyed and the tone is appropriate for the audience. \\
        \midrule
        \rowcolor{Orange!15}
        \textsc{MS-Translator Translation:} \\
        \rowcolor{Orange!15}
        \textcolor{BrickRed}{出于质量目的}，此聊天互动将关闭，请记住您可以随时回来，我们将很乐意继续为您提供帮助。 \\
        \rowcolor{Orange!15}
        \midrule
        \textsc{Analysis-based Translation:} \\
         \rowcolor{Orange!15}
        \textcolor{BrickRed}{为了保证服务质量}，此次聊天交互将会关闭，请记住您随时可以回来，我们将很乐意继续为您提供帮助。 \\
        \bottomrule
    \end{tabular}
    \caption{Instruction and example for conducting target audience analysis on a given source sentence.}
    \label{tab:target-audience-analysis}
\end{table*}

\begin{table*}[t!]
    \centering
    \begin{tabular}{p{\linewidth}}
        \toprule
        \rowcolor{Goldenrod!15}
        \textsc{Instruction:} \\
        \rowcolor{Goldenrod!15}
        Perform a Key Information Analysis on the the following Source Sentence: \\
        \rowcolor{Goldenrod!15}
        Source Sentence: \verb|{source_sentence}| \\
        \rowcolor{Goldenrod!15}
        Identify and analyze the key information in the source sentence, discussing how to maintain this information's integrity in the translation. \\
        \rowcolor{Goldenrod!15}
        Analysis: \\
        \midrule
        \rowcolor{Goldenrod!15}
        \textsc{Example:} \\
        \rowcolor{Goldenrod!15}
        Joint Aid for Dogs can be fed to all dogs from any age at the `General Support' level to maintain freedom of movement and muscle condition throughout their lives. \\
        \midrule
        \rowcolor{Goldenrod!15}
        \textsc{Answer:} \\
        \rowcolor{Goldenrod!15}
        \textcolor{BrickRed}{\textbf{In this source sentence, the key information is related to the product ``Joint Aid for Dogs'' and its usage.}} The product is specifically designed for dogs and can be given to all dogs, regardless of their age. The recommended level for usage is ``General Support,'' which suggests that the product is suitable for regular maintenance rather than treating a specific condition. The purpose of the product is to maintain the dog's freedom of movement and muscle condition throughout their lives. \\
        \rowcolor{Goldenrod!15}
        ... \\
        \rowcolor{Goldenrod!15}
        In conclusion, to maintain the integrity of the key information in the translation, it is essential to accurately convey the product name, target audience, recommended level, and purpose. This can be achieved by using appropriate terminology and language, and ensuring that the translation conveys the same level of effectiveness and suitability as the source sentence. \\
        \midrule
        \rowcolor{Goldenrod!15}
        \textsc{MS-Translator Translation:} \\
        \rowcolor{Goldenrod!15}
        \textcolor{BrickRed}{狗关节援助}可以喂给任何年龄的“一般支持”级别的所有狗，以保持其一生的活动自由和肌肉状况。 \\
        \midrule
        \rowcolor{Goldenrod!15}
        \textsc{Analysis-based Translation:} \\
        \rowcolor{Goldenrod!15}
        \textcolor{BrickRed}{Joint Aid for Dogs}可以在“一般支持”级别下喂养所有年龄段的狗，以保持它们一生中的运动自由和肌肉状况。 \\
        \bottomrule
    \end{tabular}
    \caption{Instruction and example for conducting key information analysis on a given source sentence.}
    \label{tab:key-information-analysis}
\end{table*}

\begin{table*}[t!]
    \centering
    \begin{tabular}{p{\linewidth}}
        \toprule
        \rowcolor{LimeGreen!15}
        \textsc{Instruction:} \\
        \rowcolor{LimeGreen!15}
        Perform an Intent Analysis on the the following Source Sentence: \\
        \rowcolor{LimeGreen!15}
        Source Sentence: \verb|{source_sentence}| \\
        \rowcolor{LimeGreen!15}
        Analyze the original intent of the source sentence and its impact on the translation strategy. \\
        \rowcolor{LimeGreen!15}
        Analysis: \\
        \midrule
        \rowcolor{LimeGreen!15}
        \textsc{Example:} \\
        \rowcolor{LimeGreen!15}
        Your online life and real life are merging into one seamless experience, and you need security that can keep viruses, identity theft, and other digital threats from becoming real-life headaches. \\
        \midrule
        \rowcolor{LimeGreen!15}
        \textsc{Answer:} \\
        \rowcolor{LimeGreen!15}
        The original intent of the source sentence is to emphasize the increasing integration of online and real-life experiences and the need for security measures to protect against digital threats. \textcolor{BrickRed}{\textbf{This sentence is likely promoting a product or service that offers such security.}} \\
        \rowcolor{LimeGreen!15}
        The impact on the translation strategy would be to accurately convey the message of the sentence and effectively promote the product or service. This would involve using clear and concise language to highlight the merging of online and real-life experiences, as well as the importance of security in preventing potential threats. Additionally, the translation should effectively convey the sense of urgency and potential consequences of not having proper security measures in place. It may also be beneficial to use persuasive language to encourage the audience to take action and invest in the promoted product or service. \\
        \midrule
        \rowcolor{LimeGreen!15}
        \textsc{MS-Translator Translation:} \\
        \rowcolor{LimeGreen!15}
        您的在线生活和现实生活正在融合为一种无缝体验，您需要能够防止病毒、身份盗用和其他数字威胁成为现实生活中令人头疼的问题的\textcolor{BrickRed}{安全性}。 \\
        \midrule
        \rowcolor{LimeGreen!15}
        \textsc{Analysis-based Translation:} \\
        \rowcolor{LimeGreen!15}
        您的在线生活和现实生活正在融合为一个无缝的体验，您需要\textcolor{BrickRed}{安全保障}来防止病毒、身份盗窃和其他数码威胁成为现实生活的头痛。 \\
        \bottomrule
    \end{tabular}
    \caption{Instruction and example for conducting intent analysis on a given source sentence.}
    \label{tab:intent-analysis}
\end{table*}

\begin{table*}[t!]
    \centering
    \begin{tabular}{p{\linewidth}}
        \toprule
        \rowcolor{Cerulean!15}
        \textsc{Instruction:} \\
        \rowcolor{Cerulean!15}
        Perform a Information Equivalence Analysis on the the following Source Sentence: \\
        \rowcolor{Cerulean!15}
        Source Sentence: \verb|{source_sentence}| \\
        \rowcolor{Cerulean!15}
        Analyze how to convey the same information and emotions of the source sentence in \verb|{target}|. \\
        \rowcolor{Cerulean!15}
        Analysis: \\
        \midrule
        \rowcolor{Cerulean!15}
        \textsc{Example:} \\
        \rowcolor{Cerulean!15}
        Instanced Player Housing - From tents to palaces and everything in between, players can furnish and customize their own homes and can hire NPCs to keep your place clean or purchase things from the marketplace. \\
        \midrule
        \rowcolor{Cerulean!15}
        \textsc{Answer:} \\
        \rowcolor{Cerulean!15}
        The Chinese equivalent conveys the same information and emotions as the source sentence. ``Instanced Player Housing'' is translated as ``实例化玩家住房'', which captures the concept of players having their own individual and unique homes. \textcolor{BrickRed}{\textbf{The phrase ``From tents to palaces and everything in between'' is translated as ``从帐篷到宫殿，应有尽有'', which maintains the idea of a wide range of options for players to choose from.}} The phrase "furnish and customize" is translated as ``装饰和定制'', which conveys the idea of personalizing one's home. The phrase ``hire NPCs to keep your place clean'' is translated as ``雇佣NPC来保持清洁'', which captures the practical aspect of hiring NPCs to maintain the player's home. Lastly, the phrase ``purchase things from the marketplace'' is translated as ``从市场购买物品'', which conveys the idea of being able to buy items for the player's home from a marketplace. Overall, the Chinese equivalent effectively conveys the same information and emotions as the source sentence. \\
        \midrule
        \rowcolor{Cerulean!15}
        \textsc{MS-Translator Translation:} \\
        \rowcolor{Cerulean!15}
        实例化玩家住房 - \textcolor{BrickRed}{从帐篷到宫殿以及介于两者之间的一切}，玩家可以布置和定制自己的房屋，并可以雇用NPC来保持您的地方清洁或从市场上购买东西。 \\
        \midrule
        \rowcolor{Cerulean!15}
        \textsc{Analysis-based Translation:} \\
        \rowcolor{Cerulean!15}
        实例化玩家住房 - \textcolor{BrickRed}{从帐篷到宫殿}，玩家可以装饰和定制自己的家，还可雇佣NPC保持清洁或从市场购买物品。 \\
        \bottomrule
    \end{tabular}
    \caption{Instruction and example for conducting intent analysis on a given source sentence, in which ``\texttt{\{target\}}'' refers to the target translation language.}
    \label{tab:information-equivalence-analysis}
\end{table*}

\begin{table*}[t!]
    \centering
    \begin{tabular}{p{\linewidth}}
        \toprule
        \rowcolor{NavyBlue!15}
        \textsc{Instruction:} \\
        \rowcolor{NavyBlue!15}
        Perform a Cultural Equivalence Analysis on the the following Source Sentence: \\
        \rowcolor{NavyBlue!15}
        Source Sentence: \verb|{source_sentence}| \\
        \rowcolor{NavyBlue!15}
        Discuss the differences between the source and \verb|{target}| cultures, and analyze how these differences impact the translation. \\
        \rowcolor{NavyBlue!15}
        Analysis: \\
        \midrule
        \rowcolor{NavyBlue!15}
        \textsc{Example:} \\
        \rowcolor{NavyBlue!15}
        Alfie Barbeary's all-action display capped by a captivating try had given Wasps hope but he was unexpectedly replaced at half-time and it became one-way traffic from the moment Campbell showed his class to finish in the 43rd minute. \\
        \midrule
        \rowcolor{NavyBlue!15}
        \textsc{Answer:} \\
        \rowcolor{NavyBlue!15}
        The source sentence is from a sports article about a rugby match, specifically discussing the performance of a player named Alfie Barbeary. The sentence mentions his ``all-action display'' and ``captivating try'' that gave his team, Wasps, hope. However, he was unexpectedly replaced at half-time, and the match turned in favor of the opposing team, with the moment of a player named Campbell showing his ``class'' and scoring a try in the 43rd minute. \\
        \rowcolor{NavyBlue!15}
        \textcolor{BrickRed}{\textbf{Lastly, the use of the phrase ``one-way traffic'' may not have a direct equivalent in ZH culture. This phrase is used to describe a situation in which one team is dominating the match and controlling the flow of play.}} A literal translation may not convey the same meaning and impact as it does in the source culture. \\
        \rowcolor{NavyBlue!15}
        In conclusion, the cultural differences between the source and ZH cultures present several challenges for the translation of this sentence. The translator must carefully consider the cultural context and find ways to convey the same meaning and impact in the target language. This may require adapting certain phrases or finding cultural equivalents that accurately convey the intended message. \\
        \midrule
        \rowcolor{NavyBlue!15}
        \textsc{MS-Translator Translation:} \\
        \rowcolor{NavyBlue!15}
        阿尔菲·巴比里（Alfie Barbeary）的全能表现加上一次迷人的尝试，这给了黄蜂希望，但他在中场休息时出人意料地被换下，从坎贝尔在第43分钟展示他的班级的那一刻起，\textcolor{BrickRed}{它就变成了单向交通}。 \\
        \midrule
        \rowcolor{NavyBlue!15}
        \textsc{Analysis-based Translation:} \\
        \rowcolor{NavyBlue!15}
        阿尔菲·巴伯里（Alfie Barbeary）的全力表现和引人注目的进球给了瓦斯普斯（Wasps）希望，但他在半场结束时意外被换下，从坎贝尔（Campbell）在第43分钟展现出他的实力后，\textcolor{BrickRed}{比赛局势就一边倒了}。 \\
        \rowcolor{NavyBlue!15}
        \bottomrule
    \end{tabular}
    \caption{Instruction and example for conducting cultural equivalence analysis on a given source sentence, in which ``\texttt{\{target\}}'' refers to the target translation language.}
    \label{tab:cultural-equivalence-analysis}
\end{table*}

\begin{table*}[t!]
    \centering
    \begin{tabular}{p{\linewidth}}
        \toprule
        \rowcolor{Purple!15}
        \textsc{Instruction:} \\
        \rowcolor{Purple!15}
        Perform a Text Type Analysis on the following Source Sentence: \\
        \rowcolor{Purple!15}
        Source Sentence: \verb|{source_sentence}| \\
        \rowcolor{Purple!15}
        Determine the text type of the source sentence (informative, expressive, operative) and analyze how this classification guides the translation strategy. \\
        \rowcolor{Purple!15}
        Analysis: \\
        \midrule
        \rowcolor{Purple!15}
        \textsc{Example:} \\
        \rowcolor{Purple!15}
        Please keep tapping where the images are showing, are you able to see the images and follow where to tap? \\
        \midrule
        \rowcolor{Purple!15}
        \textsc{Answer:} \\
        \rowcolor{Purple!15}
        The text type of the source sentence is operative. This type of text is used to give instructions or commands to the reader, with the goal of eliciting a specific action or response. In this case, the sentence is giving instructions on how to interact with the images being displayed. \\
        \rowcolor{Purple!15}
        ... \\
        \rowcolor{Purple!15}
        \textcolor{BrickRed}{\textbf{Secondly, the use of imperative verbs, such as ``keep tapping'' and ``follow'', is common in operative texts and should be maintained in the translation.}} These verbs convey a sense of urgency and give a clear indication of what the reader should do. \\
        \rowcolor{Purple!15}
        ... \\
        \rowcolor{Purple!15}
        Overall, the operative text type guides the translation strategy to focus on clear and direct language, maintain the tone and formality of the source sentence, and consider any cultural or contextual differences that may affect the understanding of the instructions. \\
        \midrule
        \rowcolor{Purple!15}
        \textsc{MS-Translator Translation:} \\
        \rowcolor{Purple!15}
        \textcolor{BrickRed}{请继续点击图像显示的位置}，您是否能够看到图像并点击位置？ \\
        \midrule
        \rowcolor{Purple!15}
        \textsc{Analysis-based Translation:} \\
        \rowcolor{Purple!15}
        \textcolor{BrickRed}{请在图像显示的位置持续点击}，你能看到图像并按照指示进行点击吗？ \\
        \bottomrule
    \end{tabular}
    \caption{Instruction and example for conducting text type analysis on a given source sentence.}
    \label{tab:text-type-analysis}
\end{table*}

\begin{table*}[t!]
    \centering
    \begin{tabular}{p{\linewidth}}
        \toprule
        \rowcolor{Gray!15}
        \textsc{Instruction:} \\
        \rowcolor{Gray!15}
        Source Sentence: \verb|{source_sentence}| \\
        \rowcolor{Gray!15}
        Analysis: \verb|{analysis}| \\
        \rowcolor{Gray!15}
        Please give the best translation from \verb|{source}| to \verb|{target}| based on the above analysis without explanation. \\
        \rowcolor{Gray!15}
        Best Translation: \\
        \bottomrule
    \end{tabular}
    \caption{Instruction for analysis-based translation, in which ``\texttt{\{source\}}'', ``\texttt{\{target\}}'', and ``\texttt{\{analysis\}}'' refer to the source language, target translation language, and analysis, respectively.}
    \label{tab:single-analysis}
\end{table*}

\begin{table*}[t!]
    \centering
    \begin{tabular}{p{\linewidth}}
        \toprule
        \rowcolor{Gray!15}
        \textsc{Instruction:} \\
        \rowcolor{Gray!15}
        Source Sentence: \verb|{source_sentence}| \\
        \rowcolor{Gray!15}
        Please give the best translation from \verb|{source}| to \verb|{target}| without explanation. \\
        \rowcolor{Gray!15}
        Best Translation: \\
        \bottomrule
    \end{tabular}
    \caption{Instruction for zero-shot translation, in which ``\texttt{\{source\}}'' and ``\texttt{\{target\}}'' refer to the source and target translation languages, respectively.}
    \label{tab:translate-prompt}
\end{table*}

\end{CJK}
\end{document}